\def\year{2021}\relax
\documentclass[letterpaper]{article} 
\usepackage{aaai21}  
\usepackage{times}  
\usepackage{helvet} 
\usepackage{courier}  
\usepackage[hyphens]{url}  
\usepackage{graphicx} 
\usepackage{natbib}  
\usepackage{caption} 

\usepackage{subcaption}
\usepackage{hhline}
\usepackage{pifont}
\usepackage{amsfonts, amsmath, amssymb}

\title{H-FND: Hierarchical False-Negative Denoising for Distant Supervision Relation Extraction}
\author {
    Jhih-Wei Chen\textsuperscript{\rm 1}\textsuperscript{*},
    Tsu-Jui Fu \textsuperscript{\rm 2}\textsuperscript{*},
    Chen-Kang Lee \textsuperscript{\rm 3},
    Wei-Yun Ma \textsuperscript{\rm 3} \\
}
\affiliations {
    \textsuperscript{\rm 1} UC Los Angeles,
    \textsuperscript{\rm 2} UC Santa Barbara,
    \textsuperscript{\rm 3} Academia Sinica \\
    jwchen101@g.ucla.com, tsu-juifu@ucsb.edu, 
    cklee@iis.sinica.edu.tw, ma@iis.sinica.edu.tw
}
\begin{document}

\maketitle
\begingroup\renewcommand\thefootnote{*}
\footnotetext{Equal contribution.}
\endgroup

\begin{abstract}
Although distant supervision automatically generates training data for relation 
extraction, it also introduces false-positive (FP) and false-negative (FN)
training instances to the generated datasets. Whereas both types of errors degrade the
final model performance, previous work on distant supervision denoising
focuses more on suppressing FP noise and less on resolving the FN problem. 
We here propose H-FND, a hierarchical false-negative denoising
framework for robust distant supervision relation extraction, as an FN denoising solution.
H-FND uses a hierarchical policy 
which first determines whether non-relation (NA) instances should be kept,
discarded, or revised during the training process. For those learning instances
which are to be revised, the policy further reassigns them appropriate
relations, making them better training inputs. Experiments on
SemEval-2010 and TACRED were conducted with controlled FN ratios that randomly turn the
relations of training and validation instances into negatives to generate FN
instances. 
In this setting, H-FND can revise FN instances correctly and
maintains high F1 scores even when 50\% of the instances have been turned into
negatives. Experiment on NYT10 is further conducted to
shows that H-FND is applicable in a realistic setting.
\end{abstract}

\section{Introduction}
Relation extraction~\cite{re1, re2, re3} is a core task in information
extraction. Its goal is to determine the relation between two entities in a
given sentence. For instance, given the sentence ``Jobs was born in San Francisco'', 
with head and tail entities ``Jobs'' and ``San Francisco'', the relation to be extracted 
is ``Place of Birth''. Relation extraction can be applied for many 
applications, such as question answering and knowledge graph completion.

A major difficulty with supervising relation extraction models is the cost of collecting training data, against which distant supervision (DS)~\cite{ds1, ds2} is proposed. 
DS obtains the relational facts from a knowledge base and aligns 
these facts to all sentences in the corpus to generate learning instances. 
In specific,  if a relation triple $r(h, t)$ exists in a knowledge base, then for 
a sentence $s$ which mentions both the head entity $h$ and the tail entity $t$, it 
is tagged with relation $r$ to form a learning instance $(r, h, t, s)$.

\begin{table}[t]
\footnotesize
    \centering
    \vskip+0.5cm\begin{tabular}{ccc}
        \hline
        Knowledge base & Relation & \\
        \hline
        \textbf{Steve Jobs}, \textbf{San Francisco} 
        & PoB & \\
        \hhline{===}
        Corpus & Relation & Type \\
        \hline
        \textbf{Jobs} was born in \textbf{San Francisco} & 
        PoB (\ding{51}) & TP \\
        \textbf{Jobs} moved back to \textbf{San Francisco} & 
        PoB (\ding{55}) & FP  \\
        \textbf{Manuela} was born in \textbf{New York} & 
        NA (\ding{55}) & FN \\
        
        \hline
    \end{tabular}
    \caption{Distant supervision and different types of incorrectly labeled relations. 
    The head and tail entities are shown in boldface, 
    and ``PoB'' stands for the relation ``Place of Birth''.} 
    \label{tb:tp-fp-fn}
    
    \vspace{-0.4cm}
\end{table}

Although datasets for relation extraction can be generated using distant supervision, 
they contain considerable noise~\cite{noisy}. Under distant supervision, 
there are two types of noisy instances: false positives (FP) and false negatives (FN).
Table~\ref{tb:tp-fp-fn} shows an example. The FP ``Jobs moved back to
San Francisco'' should not reflect the relation `Place of Birth'. 
Also, an FN: as there is no relation between ``Manuela'' and ``New York'' in the
knowledge base, ``Manuela was born in New York'' is wrongly labeled as a non-relation
(NA) under the closed world assumption.
Both FP and FN degrade 
model performance if they are treated as correct labels at training time.
FPs harm prediction precision, while excessive FNs
lead to low recall rates. 

In addition to   
denoising methods for learning robustly with noisy data~\cite{coteaching, cleanlab}, 
many works focus on alleviating the FP problem in DS datasets, 
including those on
pattern-based extraction~\cite{pl, arnor}, multiple-instance
learning~\cite{ds2, att, rerl}, and sentence-level denoising with adversarial
training or reinforcement learning~\cite{dsgan, rlre, rlrc}. 
However, few investigate the FN problem for distant
supervision~\cite{irmie, inference}. To the best of our knowledge, there is no
previous study on this problem for deep neural networks.

In this paper, we investigate the impact of FNs on neural-based models
and propose H-FND, a hierarchical false-negative denoising framework for robust
distant supervision. Specifically, this framework integrates a deep
reinforcement learning agent which keeps, discards, or revises probable FN
instances with a relation classifier to generate revised relations. 
In addition, to constrain the study to the FN problem and to construct
ground-truth relations to further analyze model behavior, we conduct our 
research 
on the following two human-annotated datasets: 
SemEval-2010~\cite{se} 
and TACRED~\cite{tacred}, 
with controlled FN ratios that
randomly flip relations of training/validation instances into negatives to
generate FN instances. Then, we further conduct our experiment on a distantly supervised dataset 
NYT10~\cite{nyt} and fix its positive set, to demonstrate that our framework is 
applicable for resolving FN problem in a realistic setting.
In summary, our contributions are three-fold:
\begin{itemize}
	 \item We propose a denoising framework focused on false negatives in
	 relation extraction.
	 \item We present a special transfer learning scheme for pretraining denoising agent as
	 training data is not available for this pretraining task.
	 \item We show that our method revises correctly and maintains high F1 scores
	 even under a high percentage of false negatives, and is applicable in a realistic setting. 
\end{itemize}

We organize the rest of this paper as follows: Section~2 discusses the related
work, Section~3 describes our H-FND framework, Section~4 shows the experimental
results, and Section~5 concludes.

\section{Related Work}
\citet{ds0} propose distant supervision
(DS) to automatically generate labeled data for relation classification, a new
paradigm that synthesizes positive training data by aligning a knowledge base
to an unlabeled corpus, and produces negatives with a closed-world assumption.
Although this method requires no human effort for sentence labeling, it introduces FPs
and FNs into the generated data and degrades the performance of relation extraction models.

Many previous works have attempted to solve the FP problem. Among these works, 
denoising methods that utilize reinforcement learning (RL) are the most
relevant to ours. \citet{rlrc} propose a sentence-level denoising mechanism
that trains a positive instance selector using RL, and set the RL reward to 
the prediction probability of the relation classifier. 
\citet{rlre} also utilizes RL, but in a different way. It learns a denoising
agent to redistribute FPs to NA via prediction accuracy of the classifier as the RL
reward. 

To solve the FN problem, one method is to align the KB to the corpus after
performing KB completion using inference~\cite{inference}. Although this does
  reduce the number of FNs in DS datasets,   
it helps little when the
FN relations cannot be inferred from the KB, e.g., the entities mentioned in the FN are
not in the KB. IRMIE~\cite{irmie}, another method, constructs a negative set
in a more conservative sense, in which the head or tail entities have already
participated in other relation triples in the KB. Other sentences outside the positive
and negative sets are left unlabeled (labeled as RAW in original paper) to
prevent FNs. After training on the positive and negative sets, 
positive relation triples are retrieved from the unlabeled set to expand the KB, 
after which the original DS is performed 
to improve the quality of relation extraction. The final
performance of this method depends heavily on the heuristic for constructing the
negative set, which may not be applicable for all possible relation types.

To address the FN problem in DS datasets more generally, we propose a
hierarchical denoising method to mitigate the negative effect of FNs, ensuring
a more robust relation extraction model when the presence of FN instances is
  unavoidable.   

\begin{figure*}[ht!]
\centering
    \includegraphics[width=.90\linewidth]{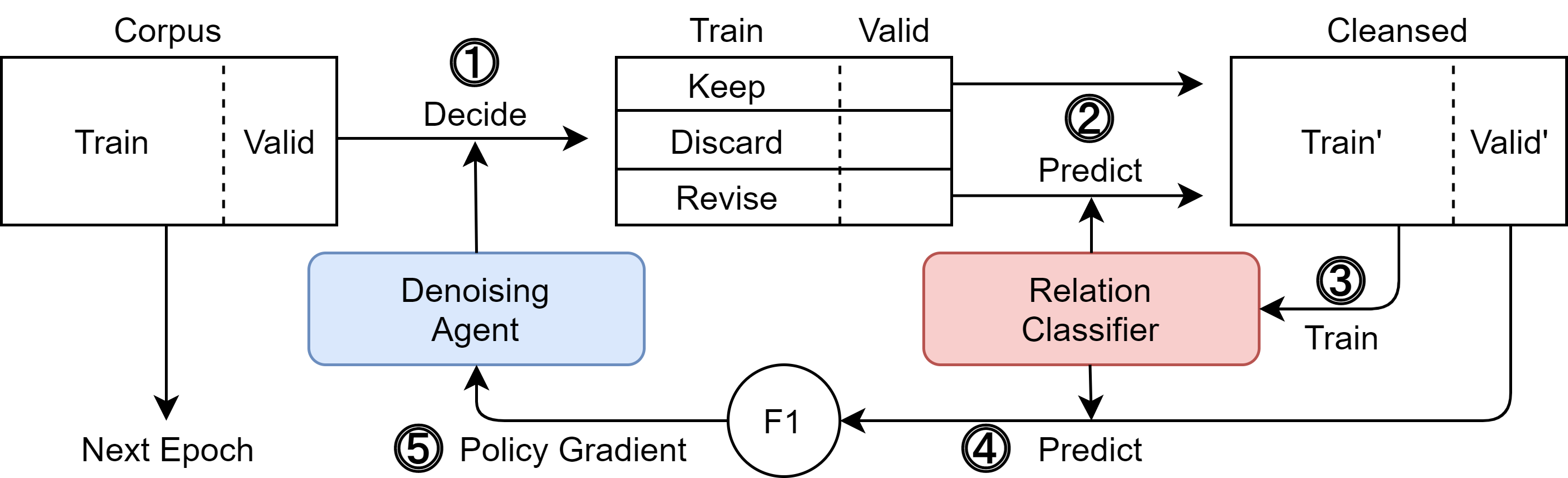}
    \caption{H-FND framework. The process in this diagram is executed per epoch.}
    \label{fig:overview}
    
\end{figure*}

\section{H-FND Framework}

We propose H-FND, a hierarchical false-negative denoising framework
that determines whether to keep, discard, or revise negative instances.
As illustrated in Fig.~\ref{fig:overview}, H-FND is composed of the
denoising agent and relation classifier modules. The denoising agent makes a
ternary decision on the action to take on each negative instance, and after discarding, the relation
classifier predicts a new relation for each to-be-revised instance to produce a
cleaned dataset.

\subsection{Convolutional Neural Network}
Convolutional neural networks (CNN) are commonly adopted for sentence-level feature extraction~\cite{kim} in language understanding tasks, such as relation extraction ~\cite{zeng_cnn, cnn}. 
PCNNs~\cite{pcnn}, a variation of CNN that applies piecewise max pooling, are also widely used for extracting sentence features~\cite{att, dsgan}. We included both as the base model in our experiments to show that our framework is base model agnostic.
In our implementation, the extracted features of a learning instance $s$ are fed into a fully connected softmax classifier to compute the final logits: 
\begin{align*}
	O(r) &= \text{softmax}(\text{FC}(\text{CNN}(s))).
\end{align*}
For detailed mathematical descriptions of CNN and PCNN, please refer to the Appendix.


\subsection{Hierarchical Denoising Policy}
The proposed hierarchical denoising policy is a framework using policy-based
reinforcement learning (RL).
Previous work utilizing RL to suppress noise from FPs~\cite{rlrc, rlre} 
can be categorized in two types of strategies: the
first decides whether to remove the input instance, and the second
decides whether to revise the input instance to be negative. Both
policies make a binary decision on each input instance, and
successfully reduce FP instances in DS datasets. 

While applicable on the FP problem, it is risky to directly apply these strategies on the FN problem. First, discarding a negative
instance even when it is 
  most likely positive   
  can result in a loss of useful learning instances.   
Second, changing a negative instance to positive 
is not enough for the training process: we must also 
know which type of positive relation to revise to. 


Therefore, we propose a hierarchical denoising policy to perform the FN
denoising in two steps. The first step, a soft policy that combines the two
above-mentioned denoising methods, is an agent that takes an action
from the action set \emph{\{Keep, Discard, Revise\}} for a negative instance $s$:
\begin{itemize}
	 \item \emph{Keep:} maintain $s$ as a negative instance for
	 training/validation;
	 \item \emph{Discard:} remove $s$ to prevent it from misleading the model;
	 \item \emph{Revise:} predict a new relation type for $s$ and treat it as
	 a positive for the following training/validation.
\end{itemize}
The policy $\pi(a|s)$ of this ternary decision is calculated based on the
sentence feature extracted from $s$ with the base model CNN encoder:
\begin{align*}
	\pi(a|s) &= \text{softmax}(\text{FC}_1(\text{CNN}_1(s)));
\end{align*}
each action $a$ has the possibility of $\pi(a|s)$ of being taken by the
denoising agent.

Then, if the negative instance $s$ is to be revised, the hierarchical policy
goes on to the second step and gives the revised relation by selecting the most
likely relation (excluding NA) predicted by the relation classifier:
\begin{align*}
	r' = \textrm{arg} \max_{r \in R \setminus \{NA\}} \text{FC}_2(\text{CNN}_2(s)) .
\end{align*}

\begin{figure*}[t]
\centering
    \includegraphics[width=.90\linewidth]{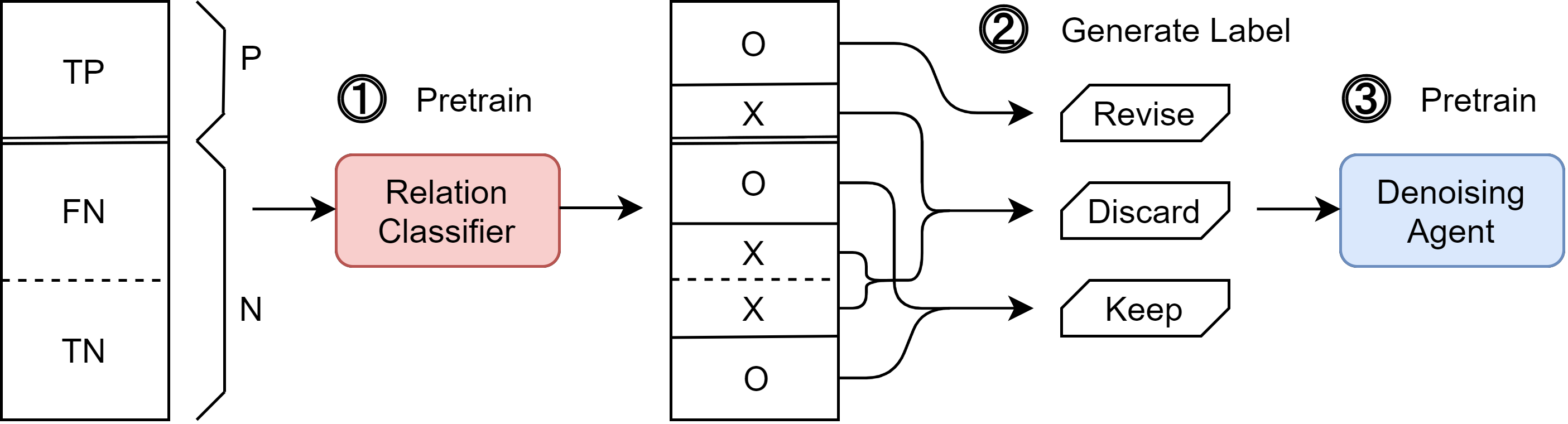}
	 \caption{A special transfer learning scheme for H-FND pretraining. Symbols ``P'' and ``N'' represent positive
	 and negative instances for relation classifier pretraining. Symbols
	 ``O'' and ``X'' indicate two sets of training instances which are correctly
	 predicted and wrongly predicted by pretrained relation classifier
	 correspondingly.}
    \label{fig:pretrain}
    
    \vspace{-0.4cm}
\end{figure*}

\subsection{Pretraining}
Supervised pretraining~\cite{rlre}, commonly used to accelerate 
RL agent training, is easily performed for the relation classifier on the original DS
dataset~\cite{coteaching}. For the denoising agent, however, there is no available
training data. Therefore, we propose a special transfer learning scheme
that utilizes the learnt knowledge in the relation classifier (source domain) to help generate action labels
for pretraining denoising agent (target domain). (See Fig.~\ref{fig:pretrain}).

First, we select the positives for which the pretrained relation classifier
correctly predicts the relation, and tag these with \emph{Revise}.
This prepares the denoising agent to identify positive
instances in the negative set in future training, and then pass these kinds
of instances to the relation classifier to predict the correct positive
relations for them. Similarly, we tag with \emph{Keep} those negatives 
  correctly predicted by the relation classifier.   
Lastly, for
instances in which the relation classifier wrongly predicts their relation, we
tag them with \emph{Discard}, encouraging the denoising agent to discard
such instances to avoid incorrect revisions.

In summary, our pretraining strategy is thus:
\begin{enumerate}
	 \item \textbf{Relation classifier pretraining}: 
	 pretrain the relation classifier (RC) directly on the original training set
	 with the categorical loss function:
	 \begin{align*}
	    ls_{\textrm{RC}} = \text{cross-entropy}(O, G),
    \end{align*}
    where $G$ represents the distantly supervised relation in the training set. Then, 
    fix the parameters of the relation classifier for the next step.
    
	 \item \textbf{Label generation}: generate labels $H$ with the
	 predictions of the relation classifier.
	 
	 \item \textbf{Denoising agent pretraining}: Supervise the denoising agent
	 (DA) with categorical loss:
    \begin{align*}
	    ls_{\textrm{DA}} = \text{cross-entropy}(\pi, H).
    \end{align*}
    
\end{enumerate}

\subsection{Co-Training}
To combine the training of the relation classifier and the denoising agent, we
propose the following co-training framework during each epoch (see
Fig.~\ref{fig:overview}):
\begin{enumerate}
	 \item \textbf{Denoising agent decision}: At the beginning of each epoch,
	 the denoising agent first executes the denoising policy on the dataset. For both
	 training and validation sets, the policy keeps, discards, or
	 revises NA instances.
	 \item \textbf{Relation classifier revision}: For instances to be
	 revised, the relation classifier generates revision relations for them.
	 Denoising yields the cleaned training and validation sets. 
	 \item \textbf{Relation classifier training}: Given the cleaned training
	 set, we train the relation classifier in a supervised fashion based on 
	 categorical loss:
     \begin{align*}
	    ls_{\textrm{RC}} = \text{cross-entropy}(O, G'),
    \end{align*}
	 where $G'$ represents the modified training set, which contains all the positives 
	 and the kept or revised negatives. 
	 Note that discarded negatives are not included in $G'$.
    
	 \item \textbf{Reward determination}: We evaluate the trained relation
	 classifier on the cleaned validation set to obtain the F1 score, which we
	 use as reward $R$ for denoising. As the validation set is cleaned
	 by the denoising policy, $R$ reflects the efficacy of the policy.

    
	 \item \textbf{Denoising policy update}: To maximize the reward $R$, we
	 adopt policy gradient~\cite{pg} to optimize the denoising agent by
	 maximizing the objective function $J(\theta)$:
	  \begin{align*}
	    J(\theta) \approx \sum \log p(a|\theta)(R-b),
    \end{align*}
	 where $\theta$ is the parameter of the denoising policy, $p(a|\theta)$
	 represents the softmax probability of the sampled determination or revision
	 step, and $b$ is the baseline which mitigates the high variance of the
	 REINFORCE algorithm~\cite{rf}. 
	 We set   $b$ to the average reward of the previous five epochs.   
    
\end{enumerate}

For each epoch, we obtain the revised set from the original
training/validation set via the denoising policy, and H-FND finds
the best denoising policy adaptively between supervised training and
reward maximization.

\begin{figure*}[ht]
\centering
\begin{subfigure}{.46\textwidth}
    \centering
    \includegraphics[width=\linewidth]{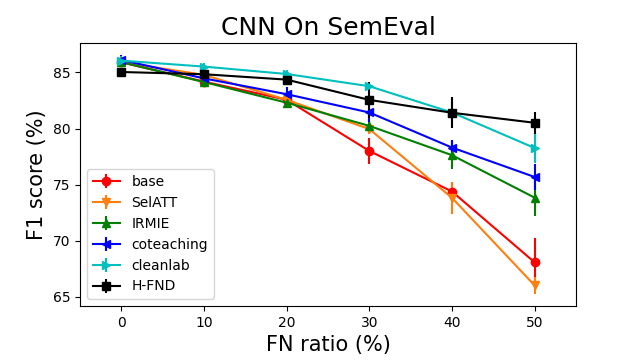}  
    \label{fig:semeval_cnn}
    \centering
    \includegraphics[width=\linewidth]{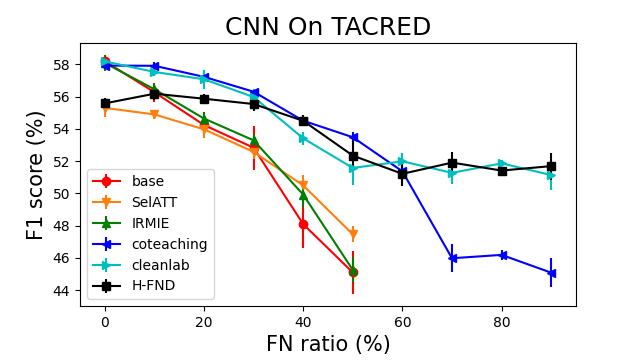}  
    \label{fig:tacred_cnn}
\end{subfigure}
\begin{subfigure}{.46\textwidth}
    \centering
    \includegraphics[width=\linewidth]{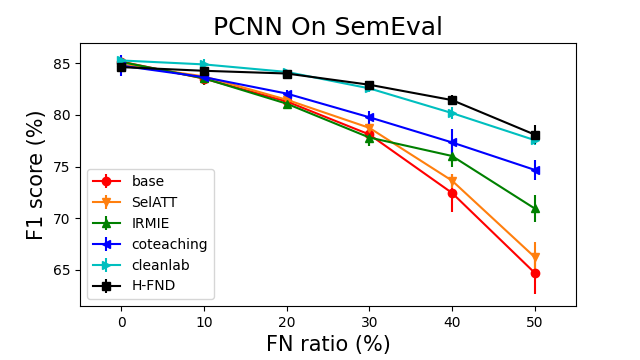}  
    \label{fig:semeval_pcnn}
    \centering
    \includegraphics[width=\linewidth]{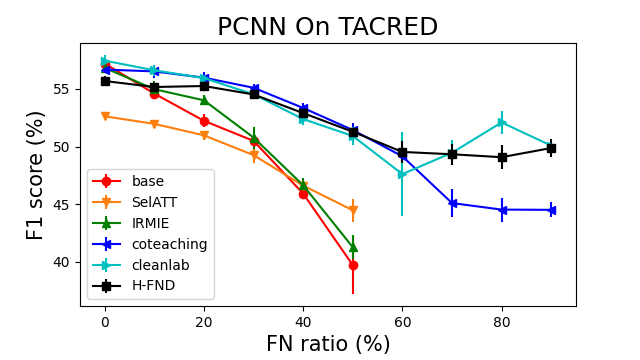}  
    \label{fig:tacred_pcnn}
\end{subfigure}
\vspace{-1\baselineskip}
\caption{CNN and PCNN results on SemEval and TACRED, where the errorbars represent the standard deviations. The denoising method cleanlab and our method H-FND perform the best, but cleanlab requires a given noise rate of data, while H-FND does not requires such information.}
\label{fig:result}
\end{figure*}

\begin{figure*}[ht]
\centering
\begin{subfigure}{.46\textwidth}
    \centering
    \includegraphics[width=\linewidth]{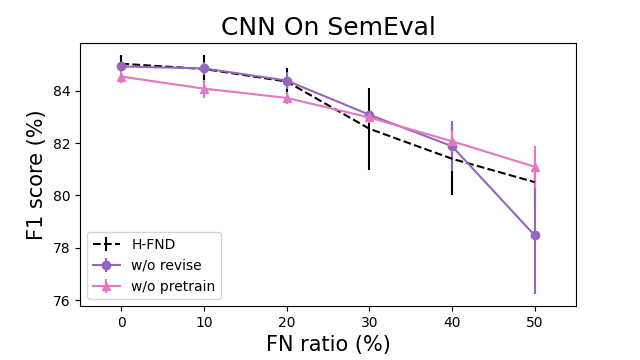}  
    \label{fig:semeval_cnn_ab}
    \centering
    \includegraphics[width=\linewidth]{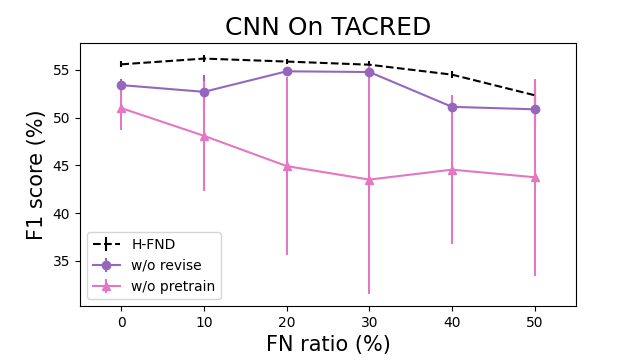}  
    \label{fig:tacred_cnn_ab}
\end{subfigure}
\begin{subfigure}{.46\textwidth}
    \centering
    \includegraphics[width=\linewidth]{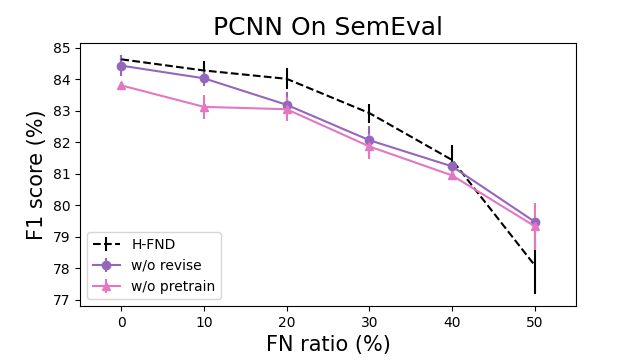}  
    \label{fig:semeval_pcnn_ab}
    \centering
    \includegraphics[width=\linewidth]{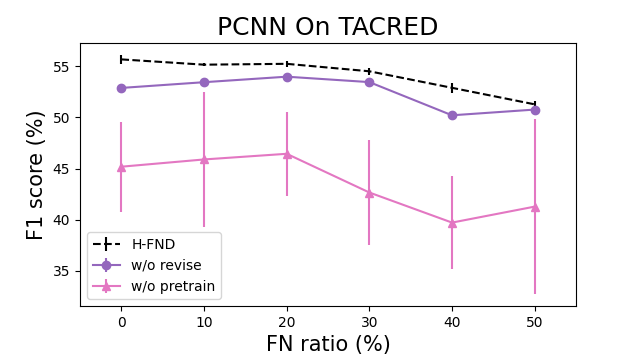}  
    \label{fig:tacred_pcnn_ab}
\end{subfigure}
\vspace{-1\baselineskip}
\caption{CNN and PCNN ablation analysis on SemEval and TACRED, where the errorbars represent the standard deviations.}
\label{fig:ablation}
\end{figure*}

\section{Experiment}
\label{sec:exp}
In order to quantify our model's performance on denoising false negatives. We first evaluated the proposed H-FND on human-annotated datasets SemEval-2010~\cite{se} and TACRED~\cite{tacred} with controlled FN ratios. Then, we evaluate H-FND on a DS dataset NYT10~\cite{nyt} to evaluate its performance in a more realistic setting. 

Table~\ref{datasets} shows the statistics of each dataset used in the experiments.
For more information of the datasets and the preprocessing procedure, please refer to Appendix.


\begin{table}
\centering
\begin{tabular}{cccc}
\hline 
\textbf{Datasets} & \#training & \#validation & \#testing \\ \hline
SemEval & 6,599 & 1,154 & 2,717 \\
\hline
TACRED & 63,782 & 20,088 & 15,509 \\
\hline
NYT10 & 477,454 & 120,318 & 194,328 \\
\hline

\end{tabular}
\caption{\label{datasets} Number of instances in each dataset}
\end{table}

\subsection{Baselines and Experiment Settings}
A simple H-FND baseline was the original CNN and PCNN relation classifier. 
To demonstrate the impact of FNs, we also included SelATT~\cite{att}, an FP 
noise resistant model. 

We further compared our H-FND framework with the following strong baselines: the
FN denoising method IRMIE~\cite{irmie} and two other general-purpose
denoising methods: co-teaching~\cite{coteaching} and cleanlab~\cite{cleanlab}.
Co-teaching is a general training method for deep neural networks to combat 
extremely noisy labels. It simultaneously maintains two networks (each with the same 
structure), each of which samples its small-loss instances with a
given overall noise rate as clean batches to its peer networks for
further training. Cleanlab is a state-of-the-art robust learning method which
directly estimates the joint distribution of noisy observed labels and latent
uncorrupted labels with a consistent estimator, filters out noisy instances
based on this joint distribution, and trains the relation classifier on the
cleaned dataset with co-teaching mentioned above. We use these denoising
methods to train the base CNN and PCNN models on our simulated FN
datasets.\footnote{The IRMIE KB was reconstructed from the positives of
the simulated FN dataset.} 

As the focus of this paper is on the FN problem, and therefore all the positives of
the simulated FN datasets are kept error-free, the H-FND framework assumes that no
positives need be changed. Hence, for a fair comparison, we kept the positive sets 
of the FN datasets unchanged for the two general-purpose denoising methods, 
preventing them from discarding error-free
positives. Also, we fix the positive set of NYT10 to evaluate the applicability of 
H-FND of resolving FN problem in a realistic setting.

In the experiments on SemEval and Tacred, every data point is the average of five independent runs. In the experiment of NYT10, some RL training is not stable, which might resulte from the excessive amount of FPs in NYT10. For a fair comparison, the included data points are the average of three best results out of five independent runs for H-FND and the baselines.
See Appendix for more detailed information on experiment and model implementation.

\subsection{Quantitative Results}
The quantitative SemEval results are shown in the upper part of
Fig.~\ref{fig:result}, including both CNN and PCNN. Under the 50\% FN ratio, 
for both the base CNN and PCNN models, with or without SelATT,
the F1 scores are heavily influenced by FN sentences: the performance drops by
nearly 20\%. ERMIE and co-teaching enhance the performance by more than 5\% and
8\% correspondingly. Except for cleanlab, H-FND denoising remains
competitive to the baselines for FN ratios from 10\% to 30\%, and significantly
wins after 30\%. Among all baselines, cleanlab's performance is the strongest
and is competitive with our approach, but as cleanlab relies on a co-teaching
model to train the relation classifier, a given  noise rate is
required. In our experiments, these are directly provided to the model. However,
in practice, the noise rate (the FN ratios in our experiment) is unknown and must
be estimated correctly, entailing extra effort. In contrast, H-FND has no such
requirement.



The quantitative results on TACRED are shown in the lower part of
Fig.~\ref{fig:result}. CNN, PCNN, and the two models with SelATT are all
vulnerable to FN instances. As IRMIE fails to exclude enough FNs from the negative set
on TACRED,\footnote{The size of the RAW set is 
  less than 10\% of the   original negative set   
under all FN ratios.} its performance is also strongly
influenced by FN instances. Although the F1 scores of H-FND are 2\% behind
co-teaching and cleanlab for FN ratios from 0\% to 20\%, it successfully
maintains its performance when the FN ratio exceeds 30\% and becomes competitive with
these two baselines. This 
is similar to the experimental results on
SemEval for FN ratios less than 30\%. Together with the fact that TACRED
has many more positives than SemEval, we increased the FN ratio
to 90\%. The result of this extended experiment shows that when the FN ratio exceeds
60\%, the F1 scores for co-teaching drop significantly, whereas H-FND maintains a
relatively high F1 score. Here, again, although cleanlab performs similar to ours with 
the
pre-defined FN ratios,\footnote{We have measured the performance of cleanlab when it was provided with a wrong FN ratio - 40\% FN ratio. Under the actual FN ratio of 80\% , its F1 scores dropped by 0.5\% for CNN and 1.8\% for PCNN.} the proposed approach needs no such information,
which better fits real-world circumstances of distant-supervised
relation classification.


\subsection{Ablation Study}
Fig.~\ref{fig:ablation} shows the result of the ablation study to justify the effectiveness of the \emph{Revise} action and pretraining strategy.
On Semeval, pretraining boosts the F1 score for the PCNN architecture for FN ratios
from 10\% to 40\%, but yields no significant difference for the other ratios. On TACRED,
however, the \emph{Revise} action and the pretraining strategy
clearly 
  yield  improved                            results.    
This improvement is substantial in particular for pretraining.
As TACRED has more
positive relation types and a much larger negative set, the FN denoising problem is
more severe than on SemEval; thus the pretraining strategy is crucial to provide
a better initial point for the denoising agent and to ensure more
stable performance.


\begin{figure*}[ht]
\centering
\begin{subfigure}{.46\textwidth}
     \centering
     \includegraphics[width=\linewidth]{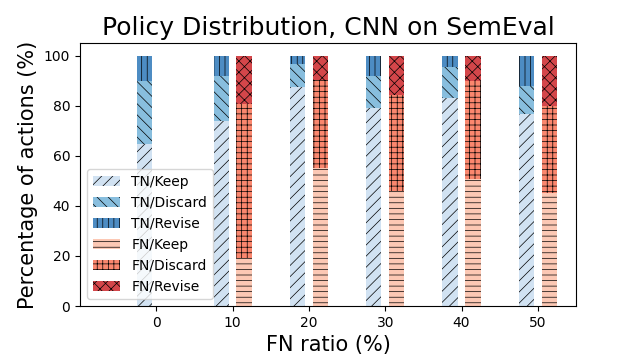}  
     \label{fig:bar_semeval_cnn}
     \centering
     \includegraphics[width=\linewidth]{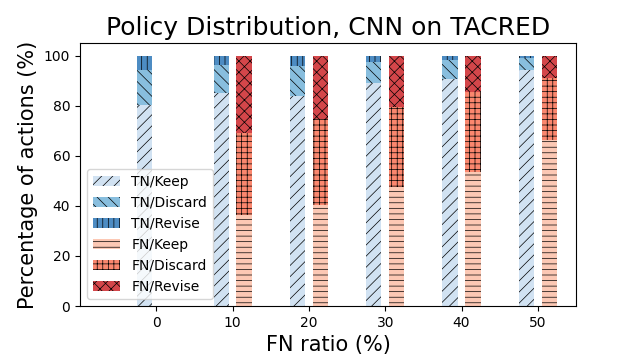}  
     \label{fig:bar_tacred_cnn}
 \end{subfigure}
 \begin{subfigure}{.46\textwidth}
     \centering
     \includegraphics[width=\linewidth]{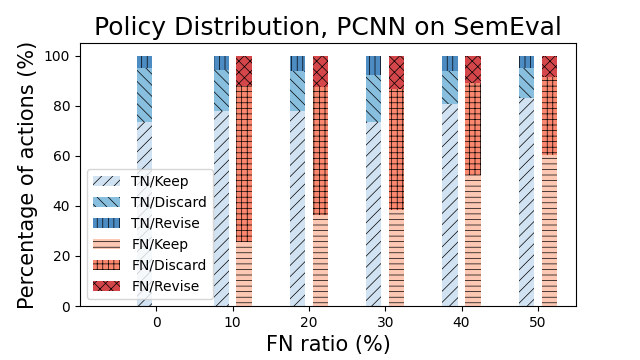} 
     \label{fig:bar_semeval_pcnn}
     \centering
     \includegraphics[width=\linewidth]{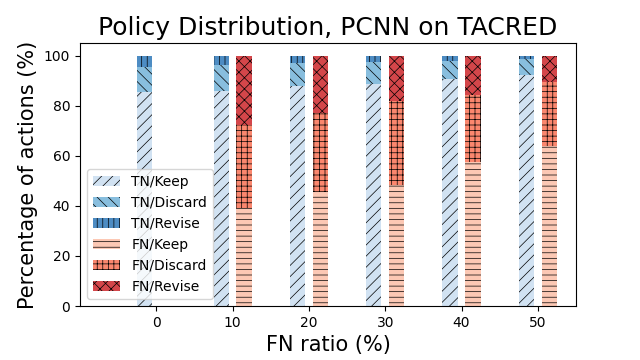}  
     \label{fig:bar_tacred_pcnn}
 \end{subfigure}
 \vspace{-1\baselineskip}
 \caption{Denoising policy distribution on true negatives and false negatives.}
 \label{fig:analysis}
 \end{figure*}
 
\begin{figure*}[ht!]
\centering
\begin{subfigure}{.46\textwidth}
    \centering
    \includegraphics[width=\linewidth]{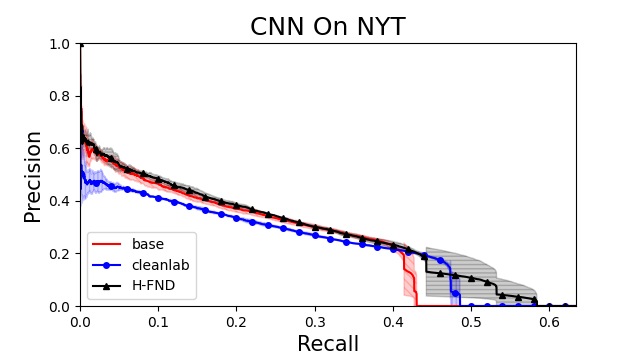}  
    \label{fig:nyt_cnn}
\end{subfigure}
\begin{subfigure}{.46\textwidth}
    \centering
    \includegraphics[width=\linewidth]{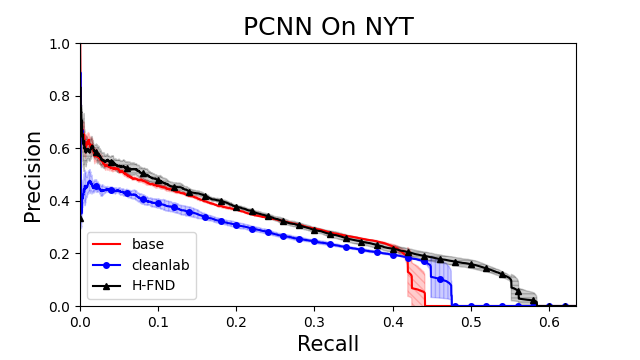}  
    \label{fig:nyt_pcnn}
\end{subfigure}
\vspace{-1\baselineskip}
\caption{Precision-recall curve on the NYT dataset. The shaded areas indicate one standard deviation. The precision rate of each algorithm run drops to zero at certain recall rate, hence the steep drops in the curves.
}\label{fig:nyt}
\end{figure*}

\subsection{Detailed Analysis}
We first analyzed the
distribution of the denoising policy for TN and FN instances in the training
set.
Figure~\ref{fig:analysis} shows the percentage of kept, discarded, or revised 
training instances. The left histogram under each filter ratio is for
TN; the right is for FN.

On SemEval, we observe that for TN instances, H-FND mainly keeps them as NA and
revises only a small portion to the wrong relation, even
under the 50\% filter ratio. For FN instances, H-FND prefers to discard or
revise them. 
This difference shows that H-FND distinguishes FN instances from TN
instances, and does not take arbitrary actions on them. 

On TACRED, the policy distribution also shares the same tendency, but
the portion of kept instances is generally larger. This is due to a higher
ratio of negative instances in TACRED. As more negative instances result in more
\emph{Keep} labels in the generated pretraining data, after pretraining, the
probability of the model taking the \emph{Keep} action is generally higher. 
It also explains that the portion of kept instances grows 
when the filter ratio is raised. 
Note that this prevents H-FND from revising too many instances at 
the beginning of co-training, making co-training more stable.


Table~\ref{rev_acc} show the
correctness of revisions on FN instances which are determined to be revised.
The accuracy is around 90\% for both CNN and PCNN architectures and for both
SemEval and TACRED. This shows that H-FND accurately corrects FN instances
once they are identified and determined to be revised in the first stage.

\begin{table}
\centering
\small
\begin{tabular}{cccccc}
\hline 
SemEval & 10\% & 20\% & 30\% & 40\% & 50\%   \\ 
\hline
CNN  & 88.81    & 88.07    & 88.72    & 86.25    & 85.94 \\
     & $\pm$ 3.55 & $\pm$ 7.80 & $\pm$ 1.04 & $\pm$ 1.53 & $\pm$ 1.46 \\
\hline
PCNN & 89.54      & 88.04      & 86.83      & 90.31      & 84.17 \\
     & $\pm$ 1.98 & $\pm$ 3.12 & $\pm$ 2.11 & $\pm$ 0.54 & $\pm$ 3.56 \\
\hline \hline
TACRED & 10\% & 20\% & 30\% & 40\% & 50\%   \\ 
\hline
CNN  & 91.43      & 90.62      & 90.89      & 91.65      & 93.39 \\
     & $\pm$ 0.72 & $\pm$ 0.99 & $\pm$ 0.63 & $\pm$ 0.99 & $\pm$ 1.79 \\
\hline
PCNN & 90.99      & 89.64      & 87.15      & 86.75      & 86.15 \\
     & $\pm$ 0.82 & $\pm$ 0.39 & $\pm$ 0.49 & $\pm$ 0.60 & $\pm$ 1.16 \\
\hline
\end{tabular}
\caption{\label{rev_acc} Revision accuracy (\%)}
\end{table}

\subsection{Results on Realistic Dataset}
Lastly, we evaluated H-FND on NYT10 to gain an understanding of our framework's performance on real DS datasets. For baselines, apart from the base model, we included cleanlab, as it is the best performing baseline in the controlled FN experiments. 
In the training set of NYT10, we conducted human evaluation on 200 randomly sampled instances and came to an estimate of 14\% of FN in the negative instances. 

We followed \citet{pcnn} and plotted the precision-recall curve to demonstrate 
the result on NYT10 (see Fig.~\ref{fig:nyt}). 
At recall rate lower than 40\% cleanlab performs slightly worse than the base 
model, while H-FND remains competitive in terms of precision.
This could be a result of inaccuracies in the estimation of FN rate in the dataset.
Since H-FND does not require a given FN rate, it is not encumbered by such 
estimation error. At higher recall rates ($>$ 50\%), H-FND retains significantly higher
precision. This result shows that H-FND is applicable for real DS datasets, 
especially when the recall rate matters.

\section{Conclusion and Future Work}
In this work, to increase the robustness of distant supervision, we present
H-FND, a hierarchical false-negative denoising framework, which keeps,
discards, or revises non-relation (NA) inputs during training and validation
phases to suppress noise from FN instances and yield a clean dataset
for relation classifiers to learn from. We also present a special transfer learning scheme for pretraining the denoising agent. 

To investigate the effects of FN instances
addressed by our approach, we generate FN instances from SemEval-2010 and
TACRED by replacing relations of instances with NA under controlled ratios. The
experimental results show that H-FND revises FN instances to
the appropriate relations and facilitates robust relation extraction.
Further experiment on NYT10 demonstrates that our framework is applicable to real world DS denoising.
This framework can be applied on tasks such as knowledge base enrichment task, where a large corpus is aligned to an incomplete knowledge base.

For large distant supervised corpora, both FP and FN instances may emerge simultaneously. Both of which should be addressed for a optimal results.
This is a challenging but very practical setting. We leave this as future work. Also, we plan to attempt other advanced relation classification approach like R-BERT~\cite{rbert} to replace CNN or PCNN in our architecture.


The source code will be released on https://github.com... 

\clearpage
\bibliography{cite}

\clearpage
\appendix
\section{Appendices}
\label{sec:appendix}
\subsection{Convolutional Neural Network}
We use a convolutional neural network (CNN)~\cite{cnn} as our base model for both the denoising agent and the relation classifier. This architecture consists of four main layers (the first three layers compose the CNN encoder):

\begin{enumerate}
	 \item \textbf{Embedding}: The embedding layer transforms a word into a
	 vector representation, which is a concatenation of a word embedding
	 $\mathit{V_w}$ and a pair of positional embedding vectors
	 $\mathit{V_{p_1}}$, $\mathit{V_{p_2}}$~\cite{att}. Word embedding
	 $\mathit{V_w}$ is a vector that represents the semantics of a word, and
	 positional embedding pair $\mathit{V_{p_1}}, \mathit{V_{p_2}}$ is two
	 vectors representing the relative distance from the current
	 word to two entities in the sentence. 

	 The final embedding vector $\mathit{V}$ of dimension $d_e$ for each word is
	 the concatenation of $\mathit{V_w}$, $\mathit{V_{p_1}}$, and
	 $\mathit{V_{p_2}}$:
	 \begin{align*}
	     \mathit{V} = \left [ \mathit{V_w} | 
	           \mathit{V_{p_1}} | \mathit{V_{p_2}}\right ].
	 \end{align*}
	 
     \item \textbf{Convolution}: The convolutional layer transforms the embedding 
	 vectors of words into local features by applying sliding filters over them. 
	 Each filter consists of a weight matrix $A_i \in \mathbb{R}^{f \times d_e}$ and 
	 a bias term $b_i \in \mathbb{R}$, to extract specific patterns in the 
	 embedding vectors. With $h$ filters of length $f$, the entry in the feature map 
	 $C_f \in \mathbb{R}^{h \times (L-f+1)}$ for the $i$-th filter at position $t$ is
	 \begin{align*}
	     [C_f]_{it} &= \sum_{j=1}^{f} \sum_{k=1}^{d_e}
	               A_{ijk} \cdot V_{t+j-1, k} + b_i,
	 \end{align*}
	 where $L$ is the length of the input sentence. To capture information expressed
	 in phrases of all lengths, we further use $n$ different lengths of filters, and 
	 concatenate all $C_f$ under filter size $f$ as the jointed feature map 
	 $C \in \mathbb{R}^{nf \times d_e}$:
	 \begin{align*}
	     C &= [C_{f_1} | C_{f_2} | \dotsb | C_{f_n}].
	 \end{align*}
    
	 \item \textbf{Max pooling}: The max pooling layer captures the most
	 significant feature into the pooling feature $P_i$ by selecting the highest
	 value in the feature map extracted by the $i$-th filter $C_i$ over all positions:
	 \begin{align*}
	     P_i = \max(C_i).
	 \end{align*}
	 
	 PCNN~\cite{pcnn} involves piecewise max pooling, which better suits the relation 
	 extraction task. It divides an input sentence into three segments based on the two selected entities, and then extracts features from all the three segments to capture fine-grained features for relation extraction. For PCNN, the extracted feature map
	 \begin{align*}
	    P_i = [\max(C_{i1}) | \max(C_{i2}) | \max(C_{i3})],
	 \end{align*}
	 where $C_{i1}$, $C_{i2}$, and $C_{i3}$ are the three feature map segments separated by the two selected entities. We also view $P$ as
	 the sentence feature, as it represents the essential features of
	 the whole sentence. 
    
	 \item \textbf{Fully connected}: The fully connected layer (FC) performs
	 relation classification based on sentence feature $P$ with softmax
	 activation over each relation. The computed logits $O(r)$ is written as
	 \begin{align*}
	    O(r) &= \text{softmax}(\text{FC}(P)) \\
	         &= \text{softmax}(\text{FC}(\text{CNN}(s))).
	 \end{align*}
\end{enumerate}

\begin{table*}[ht]
\centering
\begin{tabular}{ccccc}
\hline 
\#(\textbf{Params}) &      & \textbf{SemEval} & \textbf{TACRED} & \textbf{NYT} \\ 
\hline
RC         & CNN  & 1,318,130 & 1,347,602 & 1,388,933 \\
           & PCNN & 1,336,530 & 1,424,882 & 1,486,453 \\
\hline
RC+SelATT  & CNN  & 1,327,330 & 1,386,242 & $-$\\
           & PCNN & 1,364,130 & 1,540,802 & $-$\\
\hline
DA         & CNN  & 1,311,683 & 1,311,683 & 1,342,883 \\
           & PCNN & 1,317,203 & 1,317,203 & 1,348,403 \\
\hline
\end{tabular}
\caption{\label{n_params} Number of trainable parameters in each model.}
\end{table*}

\subsection{Datasets}

\begin{enumerate}

\item \textbf{Human-Annotated Datasets}:
SemEval-2010\footnote{http://www.kozareva.com/downloads.html} contains nine relations with an additional NA as a non-relation, and the number of instances for each relation is roughly equal.
TACRED\footnote{https://catalog.ldc.upenn.edu/LDC2018T24} is about 10 times larger than SemEval, and it has 42 relations including NA, and the number of negative instances accounts for 80\% of the entire corpus. 
For SemEval, we used 10\% of the training set for validation, and for TACRED we simply used the dev set as the validation set (see Table~\ref{datasets}).

We filtered out the training and validation instances which had
relation triples that appeared in the testing set to eliminate any overlap between 
relation triples in the training, validation, and testing sets, to simulate the held-out evaluation settings in distant supervision~\cite{ds0}.

To simulate FN conditions, we randomly filtered a ratio (10\%--50\%) of
training/validation positives into negatives. Note that the filtering process
was only for training/validation: the testing sets were well-labeled under all FN ratios. Also note that the models were not aware in advance which sentences were TN and which were FN.

\item \textbf{Distantly Supervised Dataset}:
The NYT10 dataset\footnote{http://iesl.cs.umass.edu/riedel/ecml/} uses Freebase as knowledge base for distant supervision. The relations are extracted from a December 2009 snapshot of Freebase. Four categories of Freebase relations are used: “people”, “business”, “person”, and “location”. These types of relations are chosen because they appear frequently in the newswire corpus.
All pairs of Freebase entities that are at least once mentioned in the same sentence are chosen as candidate relation instances. For consistency with previous research~\cite{att, rlrc, rlre}, we excluded five relations: \\
\textit{'/business/company/industry'}, \\
\textit{'/business/company\_shareholder/major\_shareholder\_of'}, \\
\textit{'/people/ethnicity/includes\_groups'}, \\
\textit{'/people/ethnicity/people'}, \\
\textit{'/sports/sports\_team\_location/teams'}

This results in a total of 53 relations (including none-relation, 'NA').

The corpus is chosen from a external source articles published by The New York Times between January 1, 1987 and June 19, 2007.
The Freebase relations were divided into two parts, one for training and one for testing. The former is aligned to the years 2005-2006 of the NYT corpus, the latter to the year 2007.

Note that the FN instance in Table~\ref{tb:tp-fp-fn} is a real example in NYT10. The original sentence is: ``Born in Astoria, New York on July 19, 1924, Manuela was a long term resident of East Rockaway, New York, a graduate of City College, founder and partner in BFW Management, and dedicated long term volunteer and employee of the Helen Keller Services for the Blind.'' The head and tail entities are Manuela and New York. Their relation should be \textit{`/people/person/place\_of\_birth'}, but is labeled as \textit{NA} in NYT10.
\end{enumerate}

\subsection{Implementation}
H-FND was implemented with PyTorch 1.6.0~\cite{pytorch} in python 3.6.9.
In our implementation, we used pretrained word embeddings provided by
SpaCy~\cite{spacy} as the fixed word embeddings ($d_w=300$). The positional
embedding ($d_p=50$) was randomly initialized and then trained with the
following network; therefore the overall dimension of embedding vector $d_e
= d_w + 2 d_p = 400$. In the convolutional layer, we applied four different sizes of
filters ($f \in$ [2, 3, 4, 5]) and set all of their feature sizes to $h=230$. Both
CNN and PCNN architectures were implemented. The total trainable parameters of each models are listed in table~\ref{n_params}. To prevent overfitting, we inserted 
dropout layers with a dropout rate of 0.5 before the convolutional layer and after 
the max pooling layer.

We trained H-FND using the Adam optimizer~\cite{adam}. In addition, we used mini-batches
(batch size $b=256$) only when training the relation classifier;
the prediction of the relation classifier and both the decision and policy
gradient of the denoising agent were executed per epoch.
Last, the revised result of H-FND in each epoch was used by the classifier
only in the same epoch and did not accumulate over epochs, which means that
at the beginning of each epoch, H-FND applied the denoising policy on the
original dataset but not on the revised dataset of the last epoch.

We list in Table~\ref{exp_setting} the learning rates for base CNN and PCNN relation
classifiers (RC), for RC with SelATT, and for RC with denoising agent (DA) under
pretraining and co-training phrases. The learning rate of RC is selected from \{1e-4, 3e-4, 1e-3, 3e-3, 1e-2\}, with the F1 score on the noise-free version of SemEval and TACRED as the selection criteria. Except SelATT and DA cotraining, the learning rates for the other models are the same to the learning rate of base RC. For SelATT, the learning rate is selected from \{1e-6, 3e-6, 1e-5, 3e-5, 1e-4\}, also with the F1 score on the noise-free version of the two datasets as the selection criteria. For DA cotraining, the learning rate is selected from \{1e-6, 3e-6, 1e-5, 3e-5, 1e-4\}, with the F1 score on the SemEval and TACRED under a 50\% FN ratio as the selection criteria.

All the RC of each method are trained to converge with validation-based early stopping. In specific, we train all the model for 150 epochs on SemEval and for 200 epochs on TACRED. For NYT, we trained all the odels for 30 epochs. 

The pretraining of H-FND trains the RC and DA for 5 and 20 epochs respectively. We select these pretraining periods by the criteria that the two models can achieve about 80\% performance comparing to the converged ones. By this means, we can prevent H-FND from overfitting the noisy labels~\cite{coteaching} and initialize H-FND with good parameters for co-training.

All the implemented models are trained on NVIDIA GTX 1080 Ti and Intel(R) Xeon(R) Silver 4110 CPU, with 12GN GPU memory, 128GB RAM, clock rate 2.10 GHz, and Linux as the operating system. The expected running time for each model on each dataset is listed in Table~\ref{run_time}.

\begin{table}[ht]
\centering
\begin{tabular}{cccc}
\hline \textbf{Learning rate} & \textbf{SemEval} & \textbf{TACRED} & \textbf{NYT} \\\hline
$lr_{\text{RC}}$ & 3e-3 & 3e-4 & 3e-4\\
\hline
$lr_{\text{RC, SelATT}}$ & 1e-5 & 3e-6 & $-$\\
\hline
$lr_{\text{RC, pre}}$ & 3e-3 & 3e-4 & 3e-4\\
$lr_{\text{DA, pre}}$ & 3e-3 & 3e-4 & 3e-4\\
$lr_{\text{RC, co}}$ & 3e-3 & 3e-4 & 3e-4\\
$lr_{\text{DA, co}}$ & 1e-4 & 3e-6 & 3e-6 \\
\hline
\end{tabular}
\caption{\label{exp_setting} Learning rates.}
\end{table}

\begin{table}[ht]
\centering
\begin{tabular}{cccc}
\hline 
\textbf{Runtime}& \textbf{SemEval} & \textbf{TACRED} & \textbf{NYT}\\
\hline
Base        & 0.05  & 0.63  &  3.25  \\
SelAtt      & 1.95  & 22.70 &  $-$   \\
IRMIE       & 0.05  & 0.67  &  $-$   \\
Co-teaching & 0.10  & 1.10  &  $-$   \\
Cleanlab    & 0.25  & 6.47  &  16.25 \\
H-FND       & 0.55  & 15.28 &  44.44 \\
\hline
\end{tabular}
\caption{\label{run_time} Runtimes for models training (hrs). }
\end{table}

\subsection{Performance on Validation Set}
\begin{figure*}[ht!]
\centering
\begin{subfigure}{.48\textwidth}
     \centering
     \includegraphics[width=\linewidth]{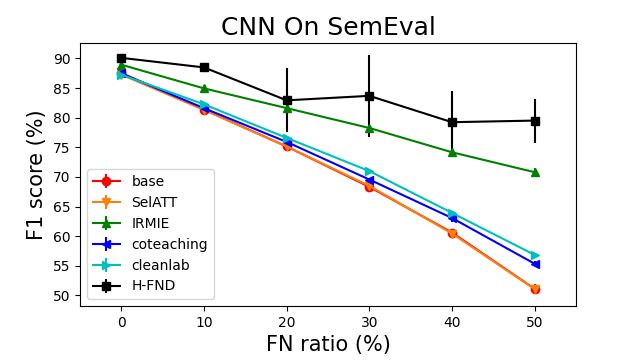}  
     \label{fig:valid_semeval_cnn}
 \end{subfigure}
 \begin{subfigure}{.48\textwidth}
     \centering
     \includegraphics[width=\linewidth]{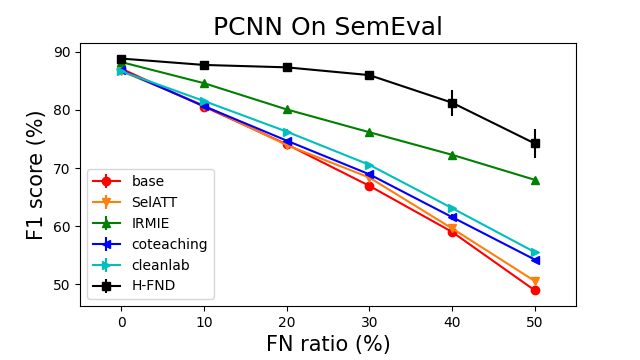}  
     \label{fig:valid_semeval_pcnn}
 \end{subfigure}
 \begin{subfigure}{.48\textwidth}
     \centering
     \includegraphics[width=\linewidth]{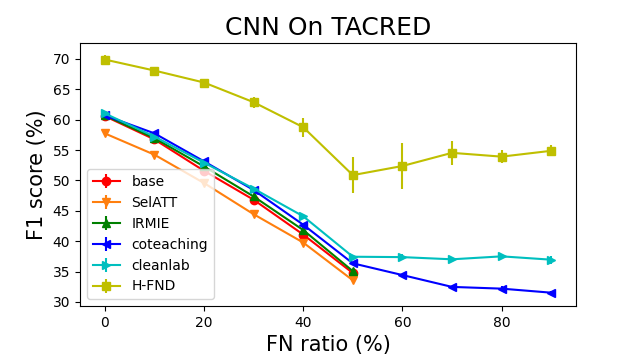}  
     \label{fig:valid_tacred_cnn}
 \end{subfigure}
 \begin{subfigure}{.48\textwidth}
     \centering
     \includegraphics[width=\linewidth]{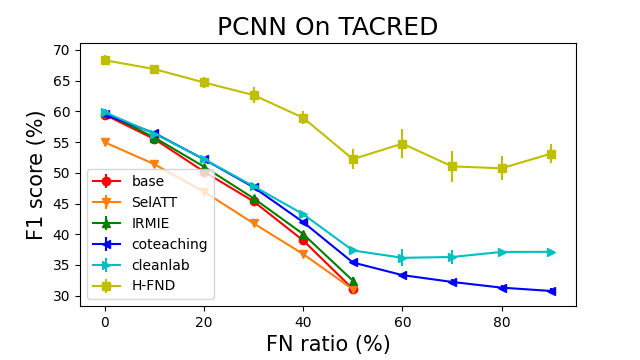}  
     \label{fig:valid_tacred_pcnn}
 \end{subfigure}
 \caption{Validation F1 scores of quantitative result, where the errorbars represent the standard deviations.}
 \label{fig:result_valid}
\end{figure*}

\begin{figure*}[ht!]
\centering
\begin{subfigure}{.48\textwidth}
     \centering
     \includegraphics[width=\linewidth]{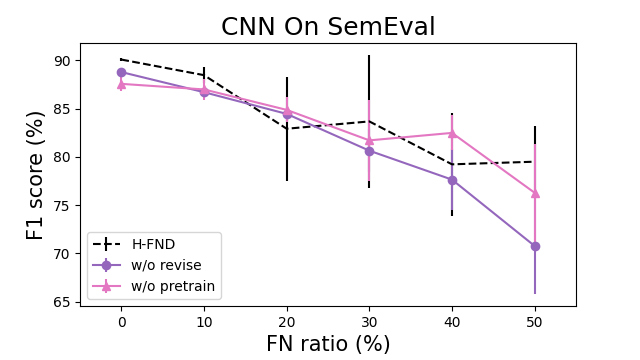}  
     \label{fig:valid_semeval_cnn_ab}
 \end{subfigure}
 \begin{subfigure}{.48\textwidth}
     \centering
     \includegraphics[width=\linewidth]{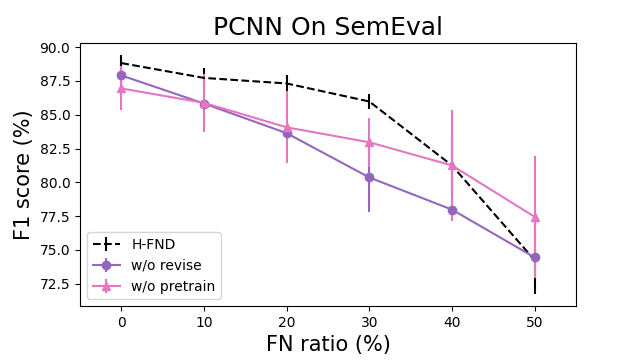}  
     \label{fig:valid_semeval_pcnn_ab}
 \end{subfigure}
 \begin{subfigure}{.48\textwidth}
     \centering
     \includegraphics[width=\linewidth]{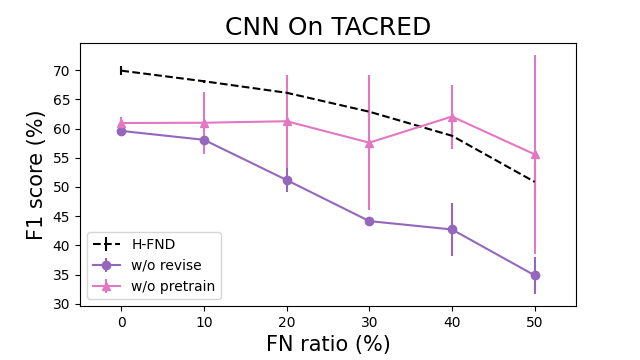}  
     \label{fig:valid_tacred_cnn_ab}
 \end{subfigure}
 \begin{subfigure}{.48\textwidth}
     \centering
     \includegraphics[width=\linewidth]{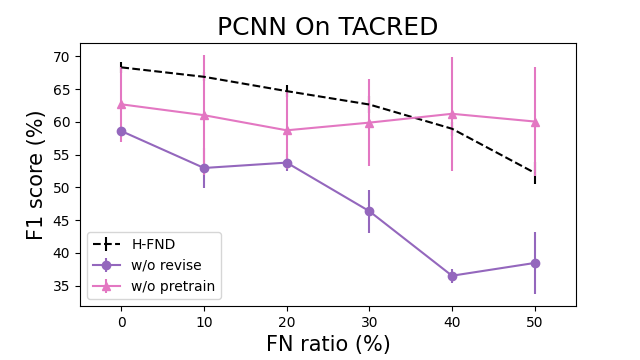}  
     \label{fig:valid_tacred_pcnn_ab}
 \end{subfigure}
 \caption{Validation F1 scores of ablation analysis, where the errorbars represent the standard deviations.}
 \label{fig:ablation_valid}
\end{figure*}

The F1 scores of each model running on validation sets of SemEval and TACRED are provided in Figure~\ref{fig:result_valid} and \ref{fig:ablation_valid}. Notice that the validation sets are noisy in our experiment, so the performance on validation sets do not fully reflect the robustness of each models. Also, in IRMIE and H-FND, the validation sets are modified, so their validation F1 scores can only be compared with their own across different FN ratios. For more accurate performance measurement, please refer to Figure~\ref{fig:result} and \ref{fig:ablation}, whose F1 scores are measured on noise-free testing sets.

\subsection{Denoising policy with Standard Deviations}
On SemEval and TACRED, the Denoising policy distributions with standard deviation are provided in Table~\ref{policy_semeval_cnn}, \ref{policy_semeval_pcnn}, \ref{policy_tacred_cnn}, and \ref{policy_tacred_pcnn}.

\begin{table*}
\centering
\small
\begin{tabular}{ccccccc}
\hline 
CNN on SemEval & 0\% & 10\% & 20\% & 30\% & 40\% & 50\%   \\ 
\hline
TN/Keep & 64.89 $\pm$ 6.55 & 74.04 $\pm$ 8.37 & 87.68 $\pm$ 9.74 & 79.17 $\pm$ 10.65 & 83.03 $\pm$ 6.65 & 76.83 $\pm$ 11.19 \\
TN/Discard & 24.88 $\pm$ 9.59 & 17.82 $\pm$ 5.55 & 8.81 $\pm$ 7.28 & 12.56 $\pm$ 6.53 & 12.34 $\pm$ 4.84 & 11.17 $\pm$ 3.15 \\
TN/Revise & 10.23 $\pm$ 5.00 & 8.14 $\pm$ 5.90 & 3.51 $\pm$ 2.57 & 8.27 $\pm$ 4.47 & 4.63 $\pm$ 1.93 & 12.00 $\pm$ 9.25 \\
FN/Keep & 0.00 $\pm$ 0.00 & 19.04 $\pm$ 5.65 & 55.03 $\pm$ 26.98 & 45.97 $\pm$ 24.85 & 50.91 $\pm$ 13.15 & 45.24 $\pm$ 11.06 \\
FN/Discard & 0.00 $\pm$ 0.00 & 61.58 $\pm$ 10.16 & 35.29 $\pm$ 22.58 & 38.33 $\pm$ 17.48 & 38.81 $\pm$ 10.05 & 34.70 $\pm$ 5.73 \\
FN/Revise & 0.00 $\pm$ 0.00 & 19.38 $\pm$ 9.77 & 9.68 $\pm$ 5.52 & 15.70 $\pm$ 7.49 & 10.28 $\pm$ 3.75 & 20.05 $\pm$ 10.99 \\
\hline
\end{tabular}
\caption{\label{policy_semeval_cnn} Denoising policy distribution for CNN on SemEval (\%).}
\end{table*}

\begin{table*}
\centering
\small
\begin{tabular}{ccccccc}
\hline 
PCNN on SemEval & 0\% & 10\% & 20\% & 30\% & 40\% & 50\%   \\ 
\hline
TN/Keep & 73.57 $\pm$ 6.19 & 77.79 $\pm$ 4.11 & 77.74 $\pm$ 3.93 & 73.61 $\pm$ 5.11 & 80.72 $\pm$ 6.08 & 82.95 $\pm$ 4.45 \\
TN/Discard & 21.63 $\pm$ 4.49 & 16.55 $\pm$ 4.71 & 16.24 $\pm$ 3.78 & 18.76 $\pm$ 5.51 & 13.25 $\pm$ 4.96 & 12.15 $\pm$ 4.21 \\
TN/Revise & 4.79 $\pm$ 2.08 & 5.67 $\pm$ 1.89 & 6.01 $\pm$ 2.16 & 7.62 $\pm$ 2.14 & 6.04 $\pm$ 2.17 & 4.91 $\pm$ 0.68 \\
FN/Keep & 0.00 $\pm$ 0.00 & 25.37 $\pm$ 5.33 & 36.46 $\pm$ 7.12 & 38.25 $\pm$ 5.81 & 52.36 $\pm$ 9.94 & 60.38 $\pm$ 8.54 \\
FN/Discard & 0.00 $\pm$ 0.00 & 62.62 $\pm$ 8.65 & 51.12 $\pm$ 8.63 & 48.45 $\pm$ 8.44 & 36.59 $\pm$ 8.87 & 31.31 $\pm$ 8.13 \\
FN/Revise & 0.00 $\pm$ 0.00 & 12.00 $\pm$ 3.51 & 12.42 $\pm$ 3.42 & 13.30 $\pm$ 3.36 & 11.05 $\pm$ 3.10 & 8.31 $\pm$ 1.42 \\
\hline
\end{tabular}
\caption{\label{policy_semeval_pcnn} Denoising policy distribution for PCNN on SemEval  (\%).}
\end{table*}

\begin{table*}
\centering
\small
\begin{tabular}{ccccccc}
\hline 
CNN on TACRED & 0\% & 10\% & 20\% & 30\% & 40\% & 50\%   \\ 
\hline
TN/Keep & 80.48 $\pm$ 3.49 & 85.18 $\pm$ 1.36 & 84.07 $\pm$ 4.66 & 89.14 $\pm$ 1.38 & 90.81 $\pm$ 1.95 & 94.11 $\pm$ 2.23 \\
TN/Discard & 13.99 $\pm$ 2.76 & 10.99 $\pm$ 1.16 & 11.78 $\pm$ 2.36 & 8.50 $\pm$ 1.18 & 7.60 $\pm$ 1.61 & 5.00 $\pm$ 1.92 \\
TN/Revise & 5.54 $\pm$ 0.84 & 3.82 $\pm$ 0.32 & 4.15 $\pm$ 2.44 & 2.35 $\pm$ 0.37 & 1.59 $\pm$ 0.37 & 0.90 $\pm$ 0.35 \\
FN/Keep & 0.00 $\pm$ 0.00 & 36.53 $\pm$ 2.21 & 40.36 $\pm$ 1.88 & 47.42 $\pm$ 3.88 & 53.60 $\pm$ 4.16 & 66.31 $\pm$ 7.85 \\
FN/Discard & 0.00 $\pm$ 0.00 & 32.40 $\pm$ 3.08 & 34.23 $\pm$ 3.38 & 32.12 $\pm$ 3.42 & 31.86 $\pm$ 2.45 & 24.60 $\pm$ 5.73 \\
FN/Revise & 0.00 $\pm$ 0.00 & 31.07 $\pm$ 1.70 & 25.42 $\pm$ 1.79 & 20.46 $\pm$ 1.86 & 14.54 $\pm$ 1.83 & 9.09 $\pm$ 2.31 \\
\hline
\end{tabular}
\caption{\label{policy_tacred_cnn} Denoising policy distribution for CNN on TACRED (\%).}
\end{table*}

\begin{table*}
\centering
\small
\begin{tabular}{ccccccc}
\hline 
PCNN on TACRED & 0\% & 10\% & 20\% & 30\% & 40\% & 50\%   \\ 
\hline
TN/Keep & 85.31 $\pm$ 0.45 & 85.91 $\pm$ 3.20 & 88.05 $\pm$ 2.73 & 88.60 $\pm$ 2.54 & 90.63 $\pm$ 2.23 & 92.46 $\pm$ 1.57 \\
TN/Discard & 10.30 $\pm$ 0.53 & 10.55 $\pm$ 2.97 & 9.10 $\pm$ 2.39 & 9.01 $\pm$ 2.07 & 7.41 $\pm$ 2.00 & 6.27 $\pm$ 1.40 \\
TN/Revise & 4.39 $\pm$ 0.34 & 3.54 $\pm$ 0.32 & 2.85 $\pm$ 0.53 & 2.39 $\pm$ 0.70 & 1.96 $\pm$ 0.31 & 1.26 $\pm$ 0.25 \\
FN/Keep & 0.00 $\pm$ 0.00 & 39.10 $\pm$ 4.23 & 45.62 $\pm$ 4.58 & 48.53 $\pm$ 6.01 & 57.50 $\pm$ 3.90 & 64.01 $\pm$ 4.44 \\
FN/Discard & 0.00 $\pm$ 0.00 & 33.03 $\pm$ 5.09 & 31.68 $\pm$ 3.64 & 33.32 $\pm$ 4.97 & 26.67 $\pm$ 3.58 & 25.55 $\pm$ 3.42 \\
FN/Revise & 0.00 $\pm$ 0.00 & 27.87 $\pm$ 1.88 & 22.70 $\pm$ 1.86 & 18.15 $\pm$ 2.82 & 15.83 $\pm$ 1.44 & 10.45 $\pm$ 1.59 \\
\hline
\end{tabular}
\caption{\label{policy_tacred_pcnn} Denoising policy distribution for PCNN on TACRED (\%).}
\end{table*}

\end{document}